\documentclass[11pt,letterpaper]{article}
\usepackage{emnlp2017}
\usepackage{times}
\usepackage{latexsym}
\usepackage{url}
\usepackage{graphicx}
\usepackage{epsfig}
\usepackage{epstopdf}
\usepackage{float}
\usepackage{multirow}
\usepackage{amsmath}
\usepackage{amsfonts}
 \emnlpfinalcopy

\def\emnlppaperid{***}

\title{Syntax Aware LSTM Model for Chinese Semantic Role Labeling}

\author{Feng Qian$^2$, Lei Sha$^1$, Baobao Chang$^1$, Lu-chen Liu$^2$, Ming Zhang$^2$  \\
  $^1$ Key Laboratory of Computational Linguistics, Ministry of Education, Peking University \\
  $^2$ Institute of Network Computing and Information Systems, Peking University \\
   \tt \{nickqian, shalei, chbb\}@pku.edu.cn\\
   \tt \{liuluchen292, mzhang\underline{\phantom{a}}cs\}@pku.edu.cn\\
 }

\date{}

\begin{document}

% \vspace{-20cm}

\maketitle

% \vspace{-30cm}

% \vspace{-20cm}
\begin{abstract}

As for semantic role labeling (SRL) task, when it comes to utilizing parsing information, both traditional methods and recent recurrent neural network (RNN) based methods use the feature engineering way. In this paper, we propose Syntax Aware Long Short Time Memory(SA-LSTM). The structure of SA-LSTM modifies according to dependency parsing information in order to model parsing information directly in an architecture engineering way instead of feature engineering way. We experimentally demonstrate that SA-LSTM gains more improvement from the model architecture. Furthermore, SA-LSTM outperforms the state-of-the-art on CPB 1.0 significantly according to Student t-test ($p<0.05$).

\end{abstract}

%###########################

\section{Introduction}

\begin{figure}[htb]
\centering
\includegraphics[width=8cm]{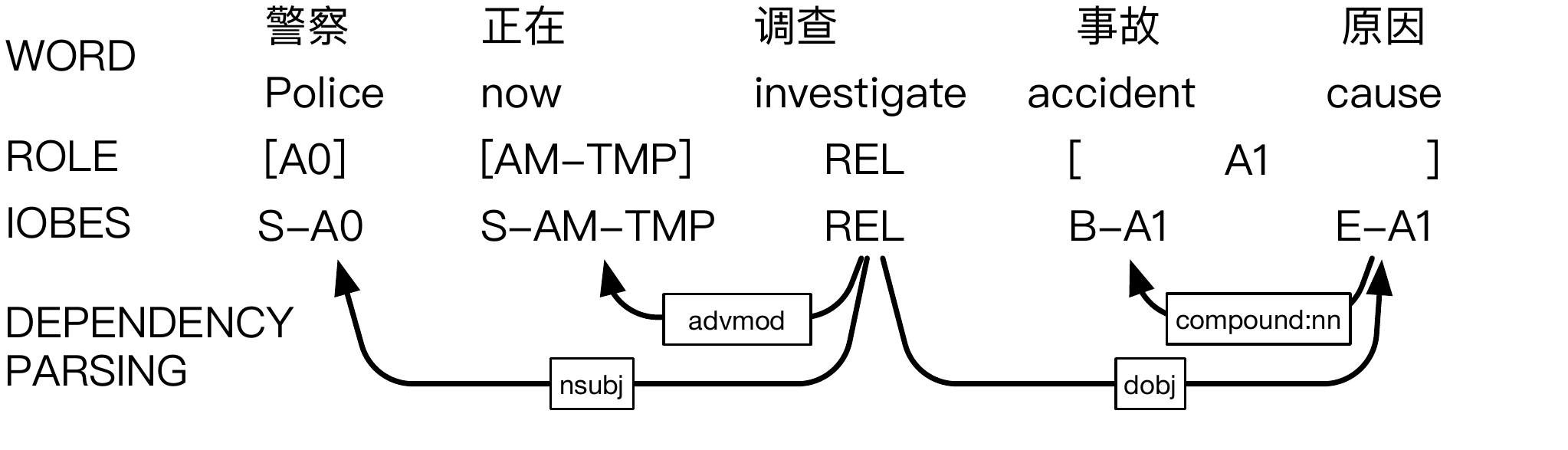}
\caption{A sentence from CPB with semantic role label and dependency parsing information} \label{fig:srl}
\end{figure}

The task of SRL is to recognize arguments of a given predicate in a sentence and assign semantic role labels. Since SRL can give a lot of semantic information, and can help in sentence understanding, a lot of NLP works such as machine translation\cite{xiong2012modeling, aziz2011shallow} use SRL information. Figure~\ref {fig:srl} shows an example of SRL task from Chinese Proposition Bank $1.0$(CPB $1.0$)\cite{xue2003annotating}.

Traditional methods on SRL use statistical classifiers such as CRF, MaxEntropy and SVM \cite{sun2004shallow, xue2008labeling,ding2008improving, ding2009word, sun2010improving} to do classification according to manually designed features.

Recent works based on recurrent neural network \cite{collobert2008unified, zhou2015end, wang2015chinese} extract features automatically, and outperform traditional methods significantly. However, RNN methods treat language as sequence data, so most of them fail to take tree structured parsing information into account, which is considered important for SRL task \cite{xue2008labeling, punyakanok2008importance, pradhan2005semantic}. Even though there are some RNN based works trying to utilize parsing information, they still do it in a feature-engineering way.

\begin{figure}[htb]
\centering
\includegraphics[width=8cm]{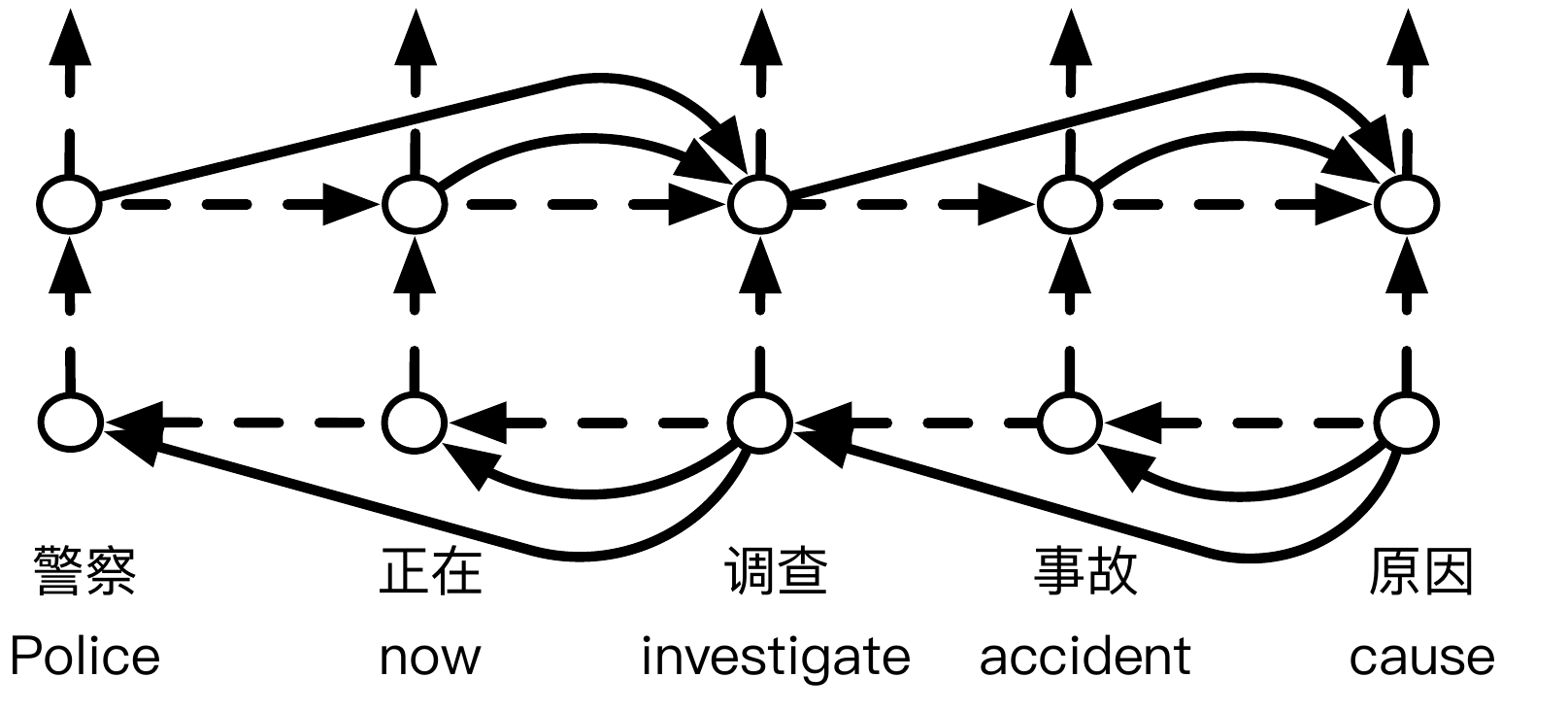}
\caption{SA-LSTM architecture: A $\bigcirc$ stands for one word. The dotted arrows stand for original neighbor connections of bi-LSTM. Solid arrows stand for dependency relationship connections. Note that though dependency parsing relationship is directed, we here treated them as undirected. We only consider whether there is a connection, and the connection type. } \label{fig:architecture}
%\vspace*{-0.5cm}
\end{figure}

We propose Syntax Aware LSTM (SA-LSTM) to directly model complex dependency parsing information in an architecture engineering way instead of feature engineering way. For example, in Figure~\ref {fig:srl}, the arrowed line stands for dependency relationship, which is rich in syntactic information. Our SA-LSTM architecture is shown in Figure~\ref {fig:architecture}. Compares to ordinary LSTM, We add additional connections between dependency related words to capture and model such rich syntactic information in architecture engineering way. Also, to take dependency relationship type into account, we also introduce trainable weights for different types of dependency relationship. The weights can be trained to indicate importance of a dependency type.

We experimentally demonstrate that SA-LSTM utilizes parsing information better than traditional feature engineering way. Furthermore, SA-LSTM reaches $79.64\% F_1$ score on CPB 1.0, outperforms the state-of-the-art significantly based on Student's t-test($p<0.05$). 

%###############################

%\vspace{-0.1cm}
\section{Syntax Aware LSTM}
%\vspace{-0.1cm}

Compares to traditional feature engineering method, RNN-LSTM alleviates the burden of manual feature design and selection. However, most RNN-LSTM based methods failed to utilize dependency parsing relationship. Based on bi-RNN-LSTM, we propose SA-LSTM which keeps all the merit points of bi-RNN-LSTM, and at the same time can model dependency parsing information directly.

%\vspace{-0.3cm}

% \begin{figure}[htb]
% \centering
% \includegraphics[width=7.8cm]{figure/architecture_2.pdf}
% \caption{This is another version for you to choose from}
% \end{figure}

%---------------------------

%\vspace{-0.1cm}
\subsection{Conventional bi-LSTM Model for SRL}

In a sentence, each word $w_t$ has a feature representation $x_t$ which is generated automatically as \cite{wang2015chinese} did. $z_t$ is feature embedding for $w_t$, calculated as followed:
 \begin{equation}
 z_t=f(W_1x_t) 
 \end{equation}

where $W_1\in \mathbb{R}^{n_1 \times n_0}$. $n_0$ is the length of word feature representation.

In a sentence, each word $w_t$ has six internal vectors, $\widetilde{C}$, $g_i$, $g_f$, $g_o$, $C_t$, and $h_t$, shown in Equation \ \ref{eq:cell}:
 \begin{equation}
 \begin{aligned}
 \widetilde{C}&=f(W_cz_t+U_ch_{t-1}+b_c) \\
 g_j&=\sigma(W_jz_t+U_jh_{t-1}+b_j)\ \ j\in\{i,f,o\} \\ \label{eq:cell}
 C_t&=g_i\odot\widetilde{C}+g_f\odot C_{t-1} \\ 
 h_t&=g_o\odot f(C_t)
 \end{aligned}
 \end{equation} 

 where $\widetilde{C}$ is the candidate value of the current cell state. $g$ are gates used to control the flow of information. $C_t$ is the current cell state. $h_t$ is hidden state of $w_t$. $W_x$ and $U_x$ are matrixs used in linear transformation:
 \begin{equation}
 \begin{aligned}
W_x,x\in\{c,i,f,o\}&\in \mathbb{R}^{n_h \times n_1}\\
U_x,x\in\{c,i,f,o\}&\in \mathbb{R}^{n_h \times n_h}\\
 \end{aligned}
 \end{equation}
As convention, $f$ stands for $tanh$ and $\sigma$ stands for $sigmoid$. $\odot$ means the element-wise multiplication. 

In order to make use of bidirectional information, the forward ${\overrightarrow{h_t}}^T$ and backward ${\overleftarrow{h_t}}^T$ are concatenated together, as shown in Equation \ref{eq:concatenate}:
  \begin{equation}
 a_t=[{\overrightarrow{h_t}}^T,{\overleftarrow{h_t}}^T] \label{eq:concatenate}
 \end{equation}

Finally, $o_t$ is the result vector with each dimension corresponding to the score of each semantic role tag, and are calculated as shown in Equation \ref{eq:output}:
 \begin{equation}
 o_t=W_3f(W_2a_t)  \label{eq:output}
 \end{equation}

where $W_2 \in \mathbb{R}^{n_3\times n_2}$, $n_2$ is $2\times h_t$, $W_3 \in \mathbb{R}^{n_4\times n_3}$ and $n_4$ is the number of tags in IOBES tagging schema.

%----------------------------------

%\vspace{-0.1cm}
\subsection{Syntax Aware LSTM Model for SRL}

\begin{figure}[htb]
\centering
\includegraphics[width=7.8cm]{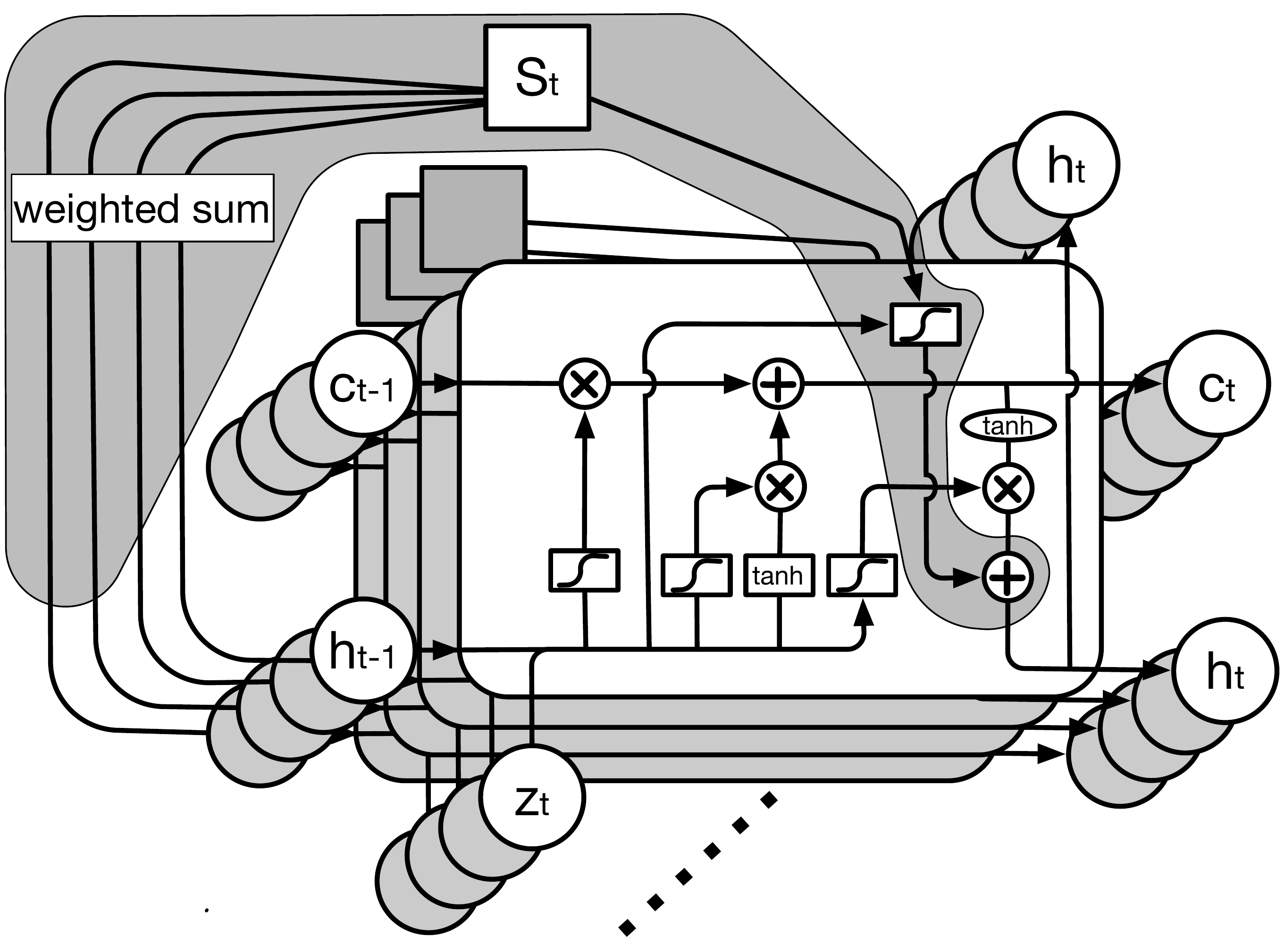}
\caption{Cell Structure of Syntax Aware LSTM} \label{fig:lstm}

\end{figure}

Structure of our SA-LSTM is shown in Figure~\ref{fig:lstm}. The most significant change we make to the original RNN-LSTM is shown in the shaded area.

$S_t$ is the syntax information input into current cell, and is calculated as shown in Equation \ref{eq: sum}:
\begin{equation}
 S_t= f(\sum_{i=0}^{t-1}\alpha\times h_i)\\ \label{eq: sum}
\end{equation}
\begin{equation}
   \alpha=
\begin{cases}
1 & \text{If there exists dependency }\\ \label{eq:alpha_1}
&\text{relationship between $w_i$ and $w_t$}\\ 
0&\text{Otherwise}
\end{cases}
\end{equation}

$S_t$ is the weighted sum of all hidden state vectors $h_i$ which come from previous words $w_i$ . Note that, $\alpha \in \{0,1\}$ indicates whether there is a dependency relationship between $w_i$ and $w_t$, only dependency related $h_i$ can be input into current cell.

We add a gate $g_s$ to constrain information from $S_t$, as shown in Equation \ref{eq:gate}. To protect the original sentence information from being diluted\cite{wu2016empirical} by $S_t$, we add $S_t$ to hidden layer vector $h_t$ instead of adding to cell state $C_t$, as shown in Equation \ref{eq:protect}:
\begin{equation}
g_s=\sigma(W_sz_t+U_sh_{t-1}+b_s) \label{eq:gate}
\end{equation}

So $h_t$ in our SA-LSTM cell is calculated as:
 \begin{equation}
 h_t=g_o\odot f(C_t)+g_s\odot S_t \label{eq:protect}
 \end{equation}

SA-LSTM changes structure by adding different connections according to dependency parsing information. In this way, we consider the whole structure of dependency tree into SA-LSTM in an architecture engineering way. 

However, by using $\alpha$ in Equation \ref{eq:alpha_1}, we do not take dependency type into account, so we further improve the way $\alpha$ is calculated from Equation\ \ref{eq:alpha_1} to Equation\ \ref{eq:alpha_2}. Each $type_{m}$ of dependency relationship is assigned a trainable weight $\alpha_{m}$. In this way SA-LSTM can model differences between types of dependency relationship.
\begin{equation}
\alpha=
\begin{cases}
\alpha_{m} & \text{If there exists $type_{m}$ dependency }\\ \label{eq:alpha_2}
&\text{relationship between $w_i$ and $w_t$}\\ 
0&\text{Otherwise}
\end{cases}
\end{equation}

\subsection{Training Criteria}
We use maximum likelihood criterion to train our model. Stochastic gradient ascent algorithm is used to optimize the parameters. Global normalization is applied.

Given a training pair $T=(x,y)$ where $T$ is the current training pair, $x$ denotes current the training sentence, and $y$ is the corresponding correct answer path. $y_t=k$ means that the $t$-th word has the $k$-th semantic role label. 
The score of $o_t$ is calculated as:
 \begin{equation}
    s(x,y,\theta)=\sum\limits_{t=1}^{N_i}o_{ty_t}
 \end{equation}

where $N_i$ is the word number of the current sentence and $\theta$ stands for all parameters.
So the log likelihood of a single sentence is
 \begin{equation}
 \begin{aligned}
 \log p(y|x,\theta)=\log\frac{exp(s(x,y,\theta))}{\sum\nolimits_{y'}exp(s(x,y',\theta))}  \\
  =s(x,y,\theta)-log\sum\nolimits_{y'}exp(s(x,y',\theta)）
 \end{aligned}
 \end{equation}
where $y'$ ranges from all valid paths of answers.

%#########################

\section{Experiment}

%-----------------------------------
\subsection{Experiment setting}

In order to compare with previous Chinese SRL works, we choose to do experiment on CPB 1.0. We also follow the same data setting as previous Chinese SRL work\cite{xue2008labeling,sun2009chinese} did. Pre-trained\footnotemark[1] word embeddings are tested on SA-LSTM and shows improvement.

\footnotetext[1]{Trained by word2vec on Chinese Gigaword Corpus}

We use Stanford Parser\cite{chen2014fast} to get dependency parsing information, which now supports Universal Dependency representation in Chinese. Note that the train set of the parser overlaps a part of our test set, so we retrained the parser to avoid overlap.

Dimension of our hyper parameters are tuned according to development set and are shown in Table\ \ref {tab:param}.\footnotemark[2]

\begin{table}[htb]\renewcommand{\arraystretch}{0.9}
\center
\begin{tabular}{|p{1.2cm}|p{0.50cm}|p{0.50cm}|p{0.50cm}|p{0.50cm}|p{1.8cm}|} 
\hline  
\textbf{Hyper Params}  &$n_1$& $n_h$& $n_2$& $n_3$ &$learning-rate$  \\
\hline
\textbf{dim}  &$200$& $100$& $200$& $100$ &$0.001$  \\
\hline
\end{tabular} 
\caption{Hyper parameter dimensions}\label{tab:param}
%\vspace{-0.6cm}
\end{table}

\footnotetext[2]{All experiment code and related files are available on request}

%-----------------------------------

\subsection{Syntax Aware LSTM Performance}

\begin{figure*}[htb]
\centering
\includegraphics[width=15cm]{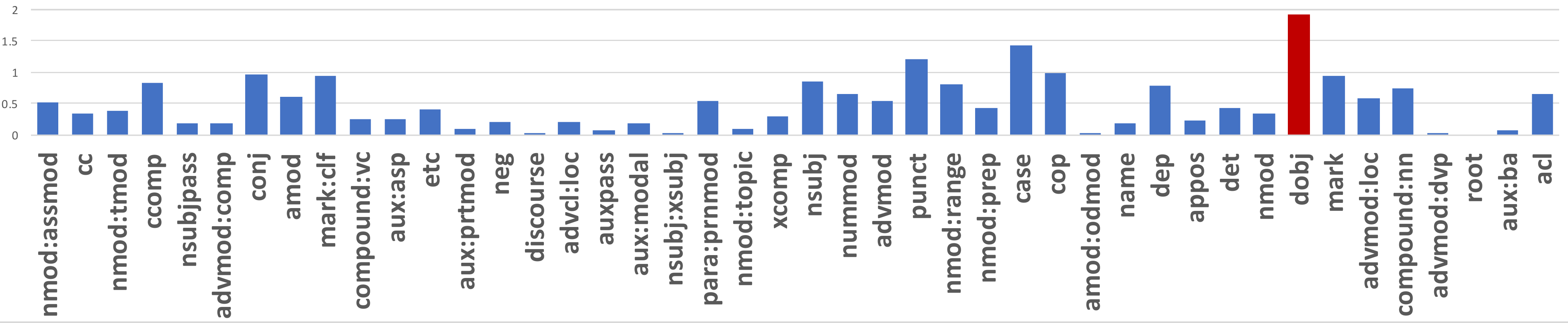}
\caption{Visualization of trained weight  $\alpha_{m}$. X axis is Universal Dependency type, Y axis is the weight.}\label{fig:weights}
\end{figure*}

\begin{table}[htb]\renewcommand{\arraystretch}{0.9}
\centering
\begin{tabular}{|p{6.48cm}|p{0.75cm}|} 
\hline  
\textbf{Method}           & \textbf{$F_1$\%}\\
\hline
Xue(2008)&71.90\\
Sun et al.(2009)&74.12\\
Yand and Zong(2014)&75.31\\
Wang et al.(2015)\textbf{(Random Initialized)}& 77.09\\
Sha et al.(2016)&\textbf{77.69}\\
\hline
Comparison Feature Engineering Way & 77.75\\
Our SA-LSTM\textbf{(Random Initialized)}& \textbf{79.56}\\
Our SA-LSTM\textbf{(Pre-trained Embedding)}& \textbf{79.64}\\
\hline
\end{tabular} 
\caption{Results comparison on CPB 1.0}\label{tab:result}
\end{table}

To prove that SA-LSTM gains more improvement from the new SA-LSTM architecture, than from the extra introduced parsing information, we design a experiment in which dependency relationship is taken into account in traditional feature engineering way.

 Given a word $w_t$, $F_t$ is the average of all dependency related $x_i$ of previous words $w_i$ , as shown in Equation \ref{eq: weighted_sum}:
\begin{equation}
 F_t= \frac{1}{T}\sum_{i=0}^{t-1} \alpha \times x_i\\ \label{eq: weighted_sum}
\end{equation}

where T is the number of dependency related words and $\alpha$ is a 0-1 variable calculated as in Equation \ref{eq:alpha_1}.

Then $F_t$ is concatenated to $x_t$ to form a new feature representation. In this way, we model dependency parsing information in a conventional feature engineering way. After that, we feed these new feature representation into ordinary bi-LSTM.

As shown in Table\ \ref{tab:result}, SA-LSTM reaches $79.56\%F_1$ score with random initialization and $79.64\%F_1$ score with pre-traind word embedding on CPB1.0 dataset. Both of them are the best $F_1$ score ever published on CPB 1.0 dataset. 

\newcite{wang2015chinese} used bi-LSTM without parsing information and got $77.09\% F_1$  score. ``comparison feature engineering method'' based on his work reaches $77.75 F_1$ score. This demonstrates the introduction of dependency parsing information has impact on SRL job.

Compared with the ``comparison feature engineering method'' shown in table~\ref{tab:result}, it is clear that SA-LSTM gain more improvement($77.75\%$ to $79.56\%$) from the architecture of SA-LSTM than from the introduction of extra dependency parsing information($77.09\%$ to $77.75\%$). Indeed, it is difficult to introduce the whole tree structure into the model using the simple feature engineering way. By building the dependency relationship directly into the structure of SA-LSTM and changing the way information flows, SA-LSTM is able to consider whole tree structure of dependency parsing information.

\subsection{Visualization of Trained Weights}

According to Equation\ \ref{eq:alpha_2}, influence from a single type of dependency relationship will be multiplied with type weight $\alpha_{m}$. When $\alpha_{m}$ is 0, the influence from this type of dependency relationship will be ignored totally. When the weight is bigger, the type of dependency relationship will have more influence on the whole system. 

As shown in Figure~\ref{fig:weights}, dependency relationship type $dobj$ receives the highest weight after training, as shown by the red bar. According to grammar knowledge, $dobj$ should be an informative relationship for SRL task, and our system give $dobj$ the most influence automatically. This example further demonstrate that the result of SA-LSTM is highly in accordance with grammar knowledge, which further validates SA-LSTM.

\section{Related works}

Semantic role labeling (SRL) was first defined by \cite{gildea2002automatic}. Early works\cite{gildea2002automatic,sun2004shallow} on SRL got promising result without large annotated SRL corpus. Xue and Palmer built the Chinese Proposition Bank\cite{xue2003annotating} to standardize Chinese SRL research.

Traditional works such as \cite{xue2005automatic, xue2008labeling, ding2009word, sun2009chinese,chen2006empirical, yang2014multi} use feature engineering methods. Traditional methods can take parsing information into account in feature engineering way, such as syntactic path feature. However, they suffer from heavy manually feature design workload, and data sparsity problem.

More recent SRL works often use neural network based methods. \newcite{collobert2008unified} proposed a convolutional neural network method for SRL. \newcite{zhou2015end} proposed bidirectional RNN-LSTM method for English SRL, and \newcite{wang2015chinese} proposed a bi-RNN-LSTM method for Chinese SRL on which our method is based. NN based methods extract features automatically and significantly outperforms traditional methods. However, most NN based methods can not utilize parsing information which is considered important for semantic related NLP tasks \cite{xue2008labeling, punyakanok2008importance, pradhan2005semantic}.

The work of \newcite{roth2016neural} and  \newcite{sha2016capturing} have the same motivation as ours, but in feature engineering way. \newcite{roth2016neural} embed dependency parsing path into feature representations using LSTM. \newcite{sha2016capturing} use dependency parsing information as feature to do argument relationships classification. In contrast, LA-LSTM utilizes parsing information in an architecture engineering way, by absorbing the parsing tree structure into SA-LSTM structure.

\section{Conclusion}

We propose Syntax Aware LSTM model for Chinese semantic role labeling. SA-LSTM is able to model dependency information directly in an architecture engineering way. We experimentally testified that SA-LSTM gains more improvement from the SA-LSTM architecture than from the input of extra dependency parsing information. We push the state-of-the-art $F_1$ to $79.64\%$, which outperforms the state-of-the-art significantly according to Student t-test($p<0.05$).

% \section*{Acknowledgments}

% Do not number the acknowledgment section.

\bibliography{emnlp2017}

\begin{thebibliography}{}
\expandafter\ifx\csname natexlab\endcsname\relax\def\natexlab#1{#1}\fi

\bibitem[{Aziz et~al.(2011)Aziz, Rios, and Specia}]{aziz2011shallow}
Wilker Aziz, Miguel Rios, and Lucia Specia. 2011.
\newblock Shallow semantic trees for smt.
\newblock In {\em Proceedings of the Sixth Workshop on Statistical Machine
  Translation\/}. Association for Computational Linguistics, pages 316--322.

\bibitem[{Chen and Manning(2014)}]{chen2014fast}
Danqi Chen and Christopher~D Manning. 2014.
\newblock A fast and accurate dependency parser using neural networks.
\newblock In {\em EMNLP\/}. pages 740--750.

\bibitem[{Chen et~al.(2006)Chen, Zhang, and Isahara}]{chen2006empirical}
Wenliang Chen, Yujie Zhang, and Hitoshi Isahara. 2006.
\newblock An empirical study of chinese chunking.
\newblock In {\em Proceedings of the COLING/ACL on Main conference poster
  sessions\/}. Association for Computational Linguistics, pages 97--104.

\bibitem[{Collobert and Weston(2008)}]{collobert2008unified}
Ronan Collobert and Jason Weston. 2008.
\newblock A unified architecture for natural language processing: Deep neural
  networks with multitask learning.
\newblock In {\em Proceedings of the 25th international conference on Machine
  learning\/}. ACM, pages 160--167.

\bibitem[{Ding and Chang(2008)}]{ding2008improving}
Weiwei Ding and Baobao Chang. 2008.
\newblock Improving chinese semantic role classification with hierarchical
  feature selection strategy.
\newblock In {\em Proceedings of the conference on empirical methods in natural
  language processing\/}. Association for Computational Linguistics, pages
  324--333.

\bibitem[{Ding and Chang(2009)}]{ding2009word}
Weiwei Ding and Baobao Chang. 2009.
\newblock Word based chinese semantic role labeling with semantic chunking.
\newblock {\em International Journal of Computer Processing Of Languages\/}
  22(02n03):133--154.

\bibitem[{Gildea and Jurafsky(2002)}]{gildea2002automatic}
Daniel Gildea and Daniel Jurafsky. 2002.
\newblock Automatic labeling of semantic roles.
\newblock {\em Computational linguistics\/} 28(3):245--288.

\bibitem[{Pradhan et~al.(2005)Pradhan, Hacioglu, Ward, Martin, and
  Jurafsky}]{pradhan2005semantic}
Sameer Pradhan, Kadri Hacioglu, Wayne Ward, James~H Martin, and Daniel
  Jurafsky. 2005.
\newblock Semantic role chunking combining complementary syntactic views.
\newblock In {\em Proceedings of the Ninth Conference on Computational Natural
  Language Learning\/}. Association for Computational Linguistics, pages
  217--220.

\bibitem[{Punyakanok et~al.(2008)Punyakanok, Roth, and
  Yih}]{punyakanok2008importance}
Vasin Punyakanok, Dan Roth, and Wen-tau Yih. 2008.
\newblock The importance of syntactic parsing and inference in semantic role
  labeling.
\newblock {\em Computational Linguistics\/} 34(2):257--287.

\bibitem[{Roth and Lapata(2016)}]{roth2016neural}
Michael Roth and Mirella Lapata. 2016.
\newblock Neural semantic role labeling with dependency path embeddings.
\newblock {\em arXiv preprint arXiv:1605.07515\/} .

\bibitem[{Sha et~al.(2016)Sha, Jiang, Li, Chang, and Sui}]{sha2016capturing}
Lei Sha, Tingsong Jiang, Sujian Li, Baobao Chang, and Zhifang Sui. 2016.
\newblock Capturing argument relationships for chinese semantic role labeling.
\newblock In {\em EMNLP\/}. pages 2011--2016.

\bibitem[{Sun and Jurafsky(2004)}]{sun2004shallow}
Honglin Sun and Daniel Jurafsky. 2004.
\newblock Shallow semantic parsing of chinese.
\newblock In {\em Proceedings of NAACL 2004\/}. pages 249--256.

\bibitem[{Sun(2010)}]{sun2010improving}
Weiwei Sun. 2010.
\newblock Improving chinese semantic role labeling with rich syntactic
  features.
\newblock In {\em Proceedings of the ACL 2010 conference short papers\/}.
  Association for Computational Linguistics, pages 168--172.

\bibitem[{Sun et~al.(2009)Sun, Sui, Wang, and Wang}]{sun2009chinese}
Weiwei Sun, Zhifang Sui, Meng Wang, and Xin Wang. 2009.
\newblock Chinese semantic role labeling with shallow parsing.
\newblock In {\em Proceedings of the 2009 Conference on Empirical Methods in
  Natural Language Processing: Volume 3-Volume 3\/}. Association for
  Computational Linguistics, pages 1475--1483.

\bibitem[{Wang et~al.(2015)Wang, Jiang, Chang, and Sui}]{wang2015chinese}
Zhen Wang, Tingsong Jiang, Baobao Chang, and Zhifang Sui. 2015.
\newblock Chinese semantic role labeling with bidirectional recurrent neural
  networks.
\newblock In {\em EMNLP\/}. pages 1626--1631.

\bibitem[{Wu et~al.(2016)Wu, Zhang, and Zong}]{wu2016empirical}
Huijia Wu, Jiajun Zhang, and Chengqing Zong. 2016.
\newblock An empirical exploration of skip connections for sequential tagging.
\newblock {\em arXiv preprint arXiv:1610.03167\/} .

\bibitem[{Xiong et~al.(2012)Xiong, Zhang, and Li}]{xiong2012modeling}
Deyi Xiong, Min Zhang, and Haizhou Li. 2012.
\newblock Modeling the translation of predicate-argument structure for smt.
\newblock In {\em Proceedings of the 50th Annual Meeting of the Association for
  Computational Linguistics: Long Papers-Volume 1\/}. Association for
  Computational Linguistics, pages 902--911.

\bibitem[{Xue(2008)}]{xue2008labeling}
Nianwen Xue. 2008.
\newblock Labeling chinese predicates with semantic roles.
\newblock {\em Computational linguistics\/} 34(2):225--255.

\bibitem[{Xue and Palmer(2003)}]{xue2003annotating}
Nianwen Xue and Martha Palmer. 2003.
\newblock Annotating the propositions in the penn chinese treebank.
\newblock In {\em Proceedings of the second SIGHAN workshop on Chinese language
  processing-Volume 17\/}. Association for Computational Linguistics, pages
  47--54.

\bibitem[{Xue and Palmer(2005)}]{xue2005automatic}
Nianwen Xue and Martha Palmer. 2005.
\newblock Automatic semantic role labeling for chinese verbs.
\newblock In {\em IJCAI\/}. Citeseer, volume~5, pages 1160--1165.

\bibitem[{Yang et~al.(2014)Yang, Zong et~al.}]{yang2014multi}
Haitong Yang, Chengqing Zong, et~al. 2014.
\newblock Multi-predicate semantic role labeling.
\newblock In {\em EMNLP\/}. pages 363--373.

\bibitem[{Zhou and Xu(2015)}]{zhou2015end}
Jie Zhou and Wei Xu. 2015.
\newblock End-to-end learning of semantic role labeling using recurrent neural
  networks.
\newblock In {\em Proceedings of the Annual Meeting of the Association for
  Computational Linguistics\/}.

\end{thebibliography}
\bibliographystyle{emnlp_natbib}

\end{document}